# Word and character segmentation directly in run-length compressed handwritten document images

Amarnath R[1], P. Nagabhushan[1, 2] and Mohammed Javed[2]

Department of Studies in Computer Science, University of Mysore.

Department of Information Technology, Indian Institute of Information Technology – Allahabad.

**Abstract –** From the literature, it is demonstrated that performing text-line segmentation directly in the run-length compressed handwritten document images significantly reduces the computational time and memory space. In this paper, we investigate the issues of word and character segmentations directly on the run-length compressed document images. Primarily, the spreads of the characters are intelligently extracted from the foreground runs of the compressed data and subsequently connected components are established. The spacing between the connected components would be larger between the adjacent words when compared to that of intra-words. With this knowledge, a threshold is empirically chosen for inter-word separation. Every connected component within a word is further analysed for character segmentation. Here, min-cut graph concept is used for separating the touching characters. Over-segmentation and under-segmentation issues are addressed by insertion and deletion operations respectively. The approach has been developed particularly for compressed handwritten English document images. However, the model has been tested on non-English document images.

## 1. Introduction

Paper documents are being used as the primary source of communication and preserving information. With the advent of imaging devices, image based documents were born which were easy to store and transmission. Generally, a digital document image occupies a large storage space and huge bandwidth for transmission. Therefore, image compression algorithms/schemes have been proposed in the literature [Khalid Sayood, 2006] to alleviate the issues of storage and transmission. In fact, modern imaging devices are equipped with inbuilt compression algorithms, which generate the digital images in the compressed format by default.

However, to perform Document Image analysis (DIA), the compressed document images have to undergo decompression and recompression stages as many times as the documents needs to be processed. Certainly, this prerequisite warrants additional buffer space and time. Hence, the efficiency gained because of compression is completely lost. To overcome this lacuna, DIA should be carried out directly in the compressed format, and so the document image compression could be viewed as an effective solution [Amarnath et al, 2017]. From this perspective, the research objective of this paper is to perform DIA directly on the compressed format of the document image

In the literature [Amarnath et al, 2018], DIA has been carried out directly in the handwritten document images for decomposing documents into line segments. Motivated from this [Amarnath et al, 2018], the proposal of this paper is to perform word and character segmentation for the applications like Optical Character Segmentation (OCR), Content Based Image Retrieval (CBIR), etc. Segmentation of handwritten text-lines into words and characters is the most challenging issues in Document Image Analysis and Recognition systems [Kumar et al, 2011; Alireza Alaei et al, 2016], where the performance of the system directly relies on the efficacy of the segmentation process [Broumandnia et al., 2008; Camastra, 2007; Fujisawa, 2010; Kherallah et al., 2008]. The task of segmentation is generally difficult because of overlapping/touching characters, distinct character shapes, spatial variation of characters, and isolated/satellite character/word fragments [Payel Rakshit et al, 2017]. As observed from the literature, most of the word and character segmentation techniques are focused on processing uncompressed document images [Kumar et al, 2011]. We could also identify some related works pertaining to the compressed machine printed document images [Javed et al, 2016]. Here, authors have attempted to traverse through base region of the text to find the word and character gaps. This may not be feasible with the compressed handwritten documents in particular. The inspiration derived from this approach is to identify the vertical gaps in all the regions such as top, middle and bottom of the text-line. In summary, in this article we aim at investigating the issue of word and character segmentation directly in the run-length compressed handwritten document images. The text-line segments that are obtained from the techniques proposed in the previous attempt will be the input to the algorithms proposed in the current model.

In this proposed work, the word segmentation approach is based on finding the inter-word gaps. Further, the spread of every character in every word segment is analysed for character segmentation. The organisation of this article is as follows: Section 2 describes the word segmentation methodology in detail. Section 3 illustrates the character segmentation model. Experimental results on compressed machine printed document images are provided in Section 4. The conclusion is provided in Section 5.

## 2. Word segmentation

In this section, word segmentation method is explained in detail. The overall idea is to find the inter-word gaps in a text-line directly in the run-length represented handwritten document image. The outline of the procedure is as follows. The first step is to identify the spread of every character in a text-line. This is achieved by screening the foreground runs and subsequently analysing its spread along the horizontal direction. In the handwritten script, most of the time, the spread of a character overlaps with its adjacent character(s). Therefore, the connectivity between such overlapping characters can be seen as a connected component. By tracing all the foreground runs, many connected components are generated for a text-line. In certain cases, multiple components are obtained for a single word. However, the appearance/frequency of components may vary depending upon the character and the word spacing. The inter-word gaps are perceivably large when compared to the intra-word gaps. Based on this clue, a threshold value is fixed empirically to distinguish words from intra-word gaps.

The detailed procedure of the model is illustrated with an example. Figure 1 shows a text-line containing 5 words.

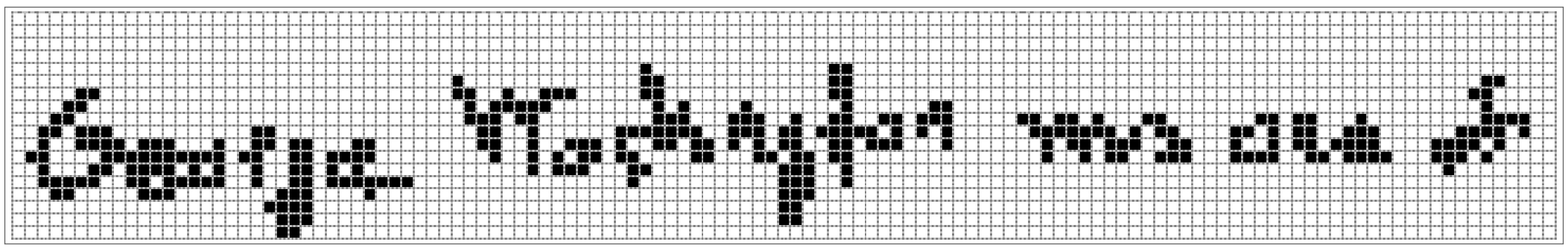

Figure 1: A sample text-line image

Figure 2 shows 11 connected components, which are indicated in red-coloured bars. These components are obtained by tracing the foreground runs.

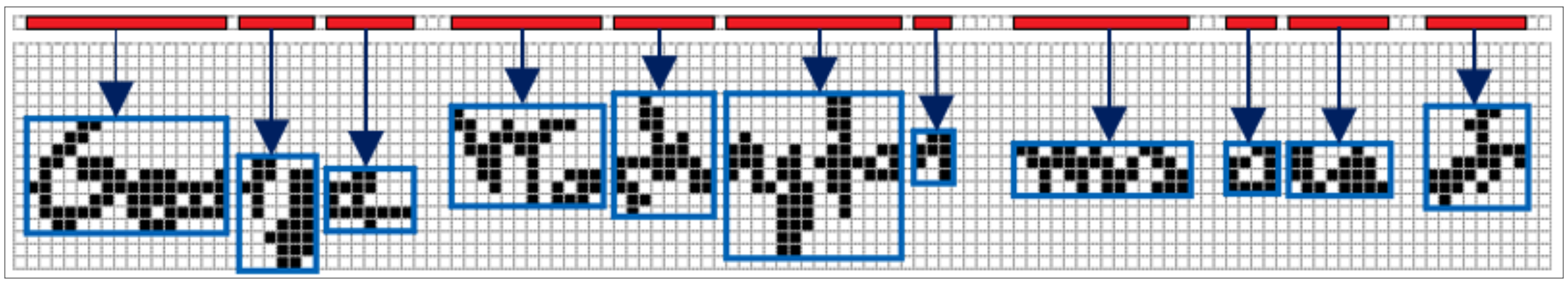

Figure 2: Connected components based on the spread of characters

Every connected component can be represented as follows

$$C_i[x_{min}, x_{max}]\ \forall i = 1..11 \qquad (1)$$

Here, $i$ represents the component number; min and max denote the spread/range of a component with respect to the x-axis.

The connected components can be obtained by applying Cumulative Run (CR). It can be formulated as

$$CR(i) = \sum_{j=1}^{n} RLE(i,j) \quad (2)$$

$$max = CR(i) \quad (3)$$

$$min = max - RLE(i,j) + 1 \quad (4)$$

Here i and n represent the row number and the total number of columns in RLE respectively.

Figure 3 illustrates the process with an example. Here we have two connected components, namely $c_1$ and $c_2$. A small gap can be spotted in column number 18. Thus, we could trace the spread of every character by using equation 1. Further, Figure 4 shows the separated components for the given example.

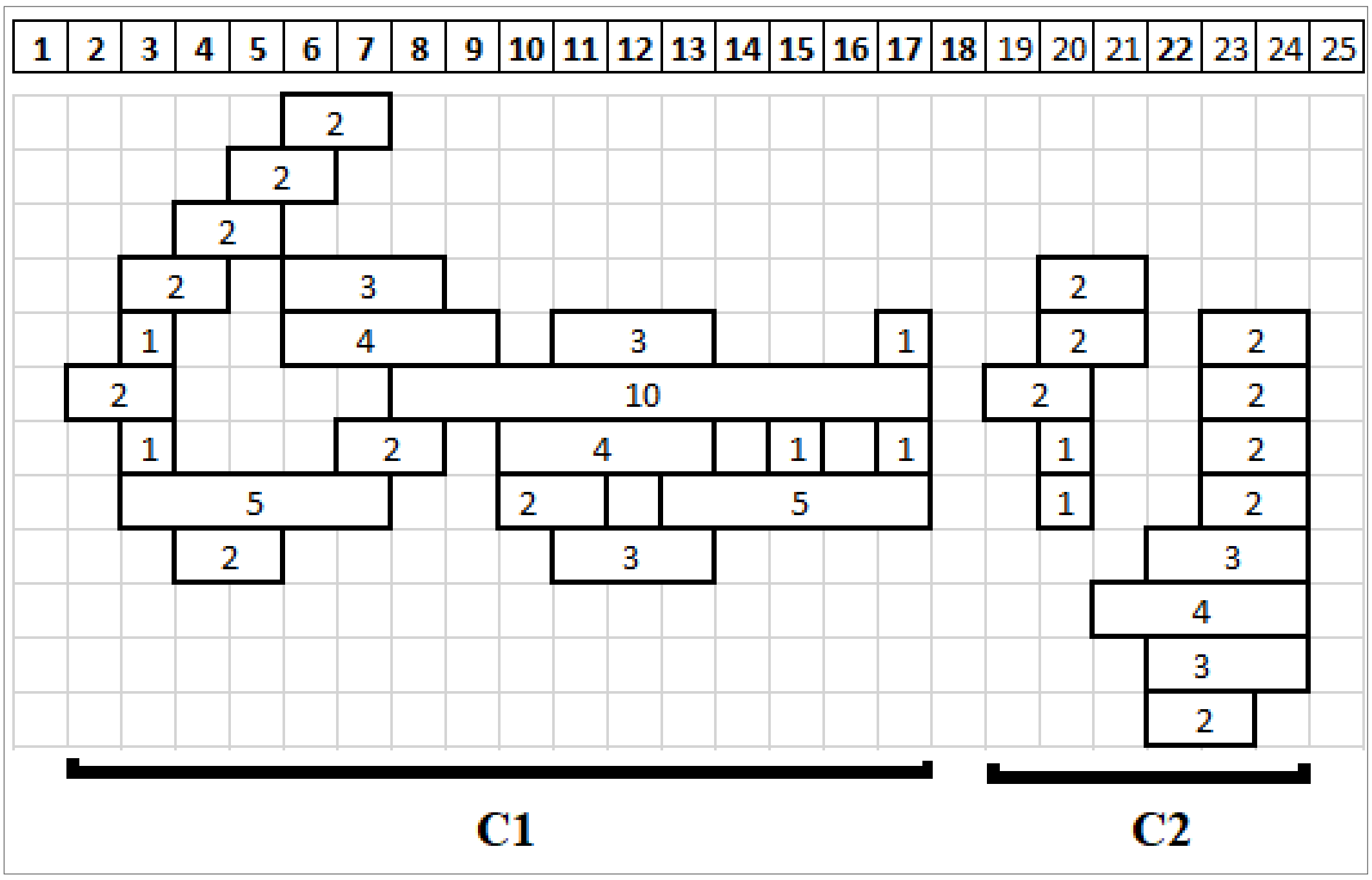

Figure 3: Spread of foreground runs for the components C1 and C2

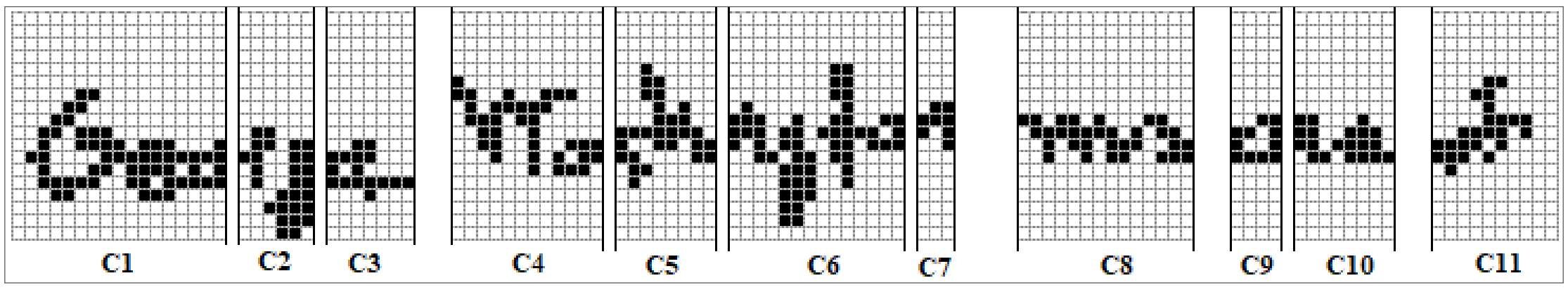

Figure 4: Spread of foreground runs for all components

Figure 5 (a) shows a sample text-line image and (b) shows the spread regions of characters respectively. Here, the inter-word gaps are found to be large when compared to the intra-word gaps.

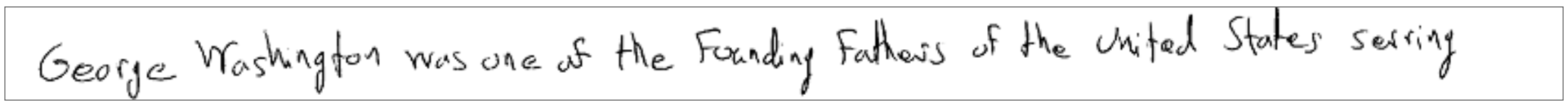

(a) A sample text-line

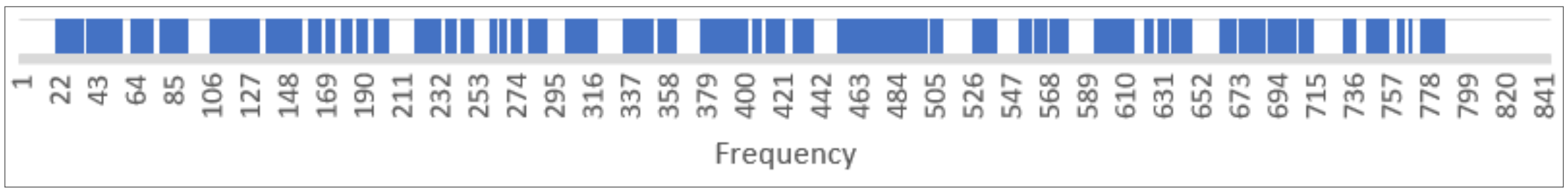

(b) Inter and Intra-word gaps

Figure 5: Sample input and the spread of characters along x-axis.

The final step is to filter out the inter-word gaps for word segmentation. This could be achieved by empirically selecting a threshold value. Table 1 shows the experimental results for the example (Figure 5). Figure 6 demonstrates the result. In this example, we could trace 13 words.

Table 1: Word segmented result for the example

| Components | Xmin | Xmax | Length |
|---|---|---|---|
| 1 | 19 | 94 | 75 |
| 2 | 105 | 204 | 99 |
| 3 | 218 | 251 | 33 |
| 4 | 260 | 292 | 32 |
| 5 | 301 | 319 | 18 |
| 6 | 333 | 363 | 30 |
| 7 | 376 | 439 | 63 |
| 8 | 452 | 510 | 58 |
| 9 | 526 | 540 | 14 |
| 10 | 552 | 579 | 27 |
| 11 | 593 | 647 | 54 |
| 12 | 662 | 715 | 53 |
| 13 | 731 | 784 | 53 |

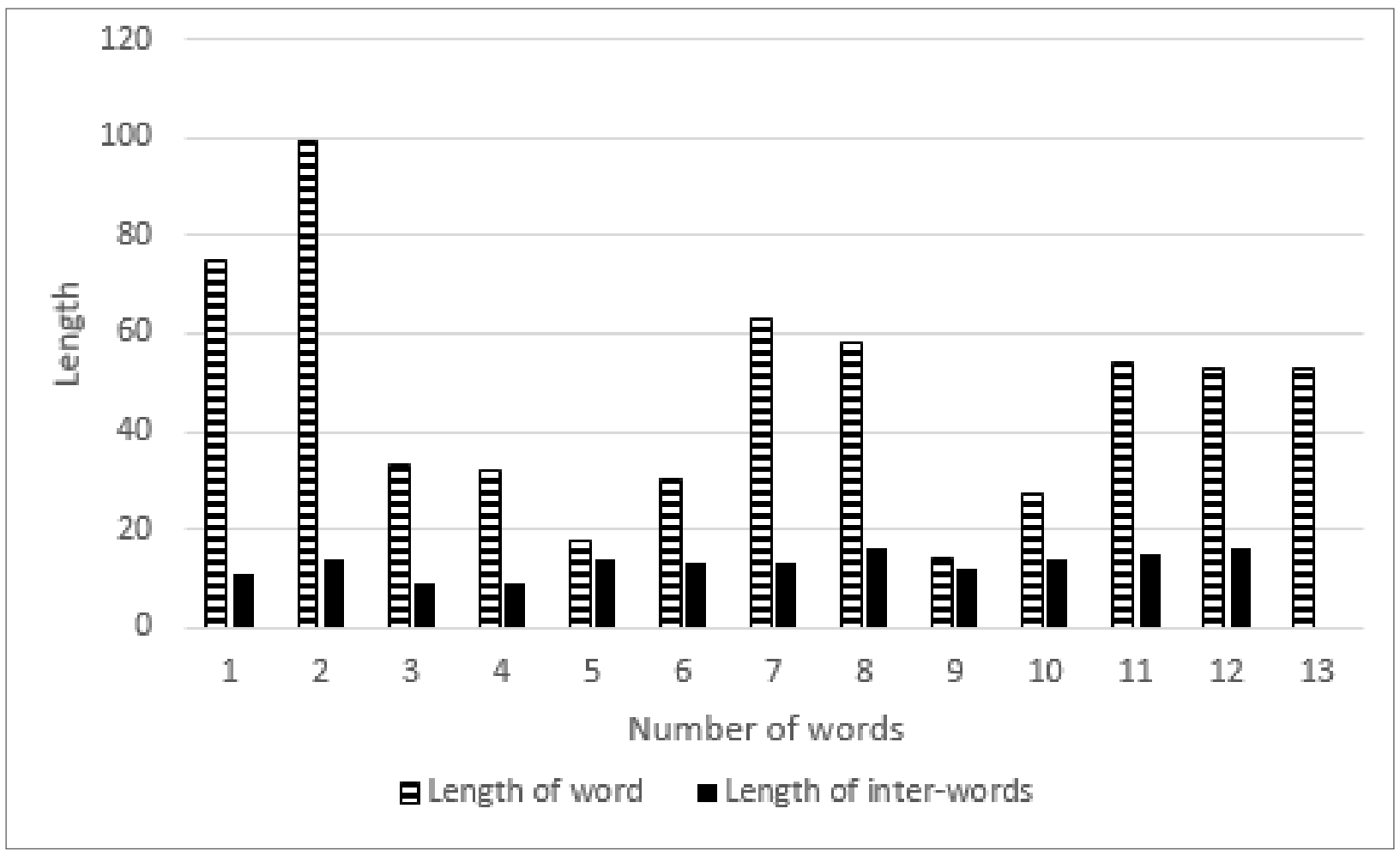

Figure 6: Lengths of the word and inter-word spacing

The final step is to find the coordinates in RLE with respect to traced spatial locations. This can be formulated as

$$x_{mid} = \frac{(x_{min} + x_{max})}{2} \tag{5}$$

$$RLE(i,j) = \begin{cases} 1, if\ CR(i,j) - RLE(i,j) < x_{mid} < CR(i,j) (using\ eq\ (2)) \\ 0, Otherwise \end{cases} \tag{6}$$

The experiments are conducted on various run-length compressed document images. Sample word segmented results are shown in Figure 7.

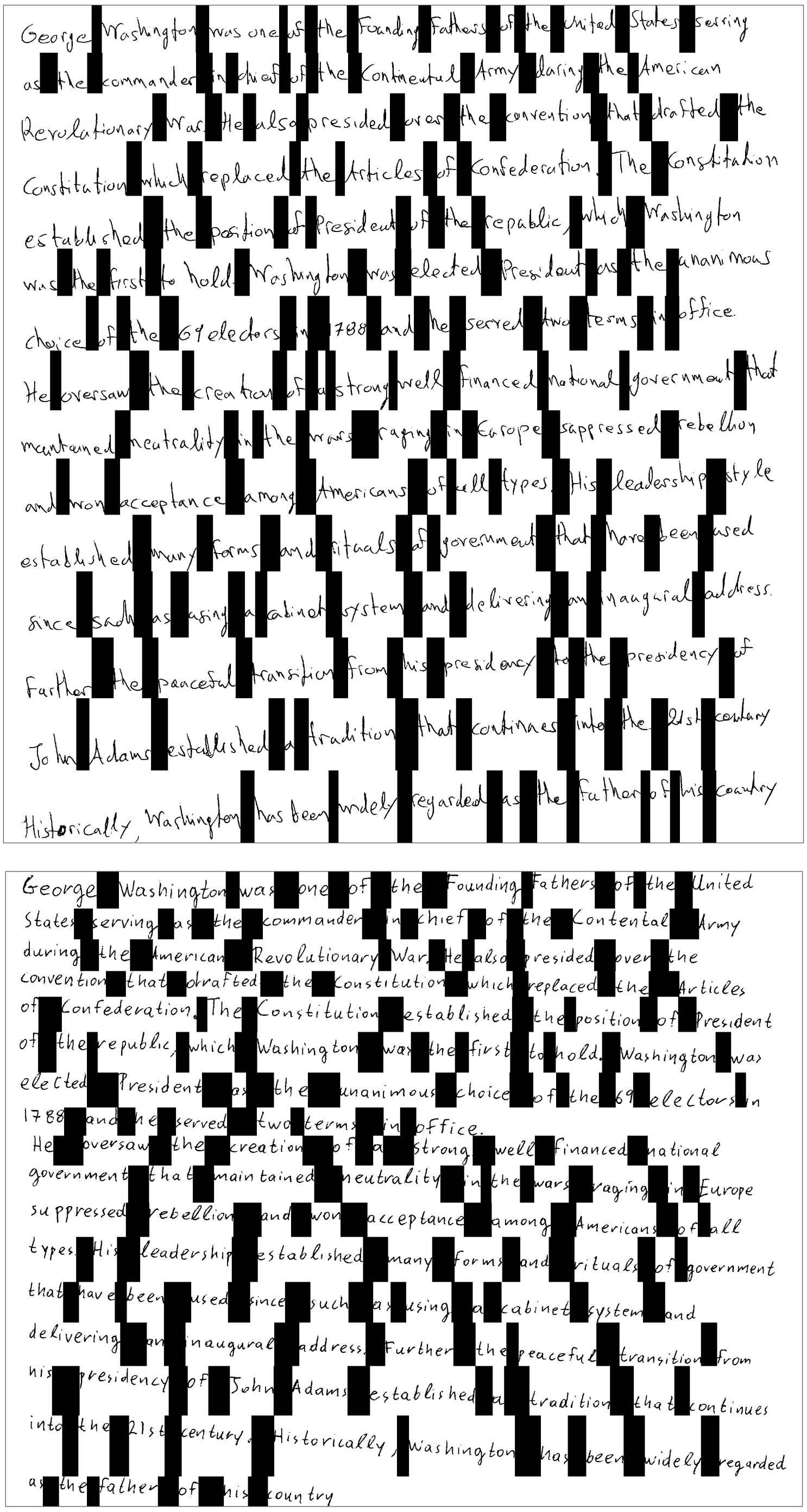

Figure 7: Results of word segmentation in the uncompressed version

## 2.1. Algorithm and Complexity Analysis

The input to the algorithm would be a run-length represented text-line image. The output would be the coordinate positions in RLE indicating word segmentation.

| | |
|---|---|
| **Algorithm:** Word segmentation | |
| ***Input:*** *Text-line segmented image in RLE format* | |
| ***Output:*** *Coordinate Position (CP) in RLE* | |
| Step 1 | Find spread of foreground runs using equations 3 and 4. |
| Step 2 | Identify the connected components with the spread information |
| Step 3 | Compute average gaps to automatically select a threshold value |
| Step 4 | Based on the threshold value, distinguish inter-words from intra-words |
| Step 5 | Identify the location in RLE using equation 6 |
| Step 6 | Stop |

The height and width of the RLE are represented as m and n respectively. The time complexity of the word segmentation algorithm takes $O\ (m \times n)$ in its worst-case scenario. If the same approach is employed in pixel domain, the best-case scenario would be $O(h \times w\ )$, where h and w represent the height and width of the uncompressed image. Figure 8 shows the comparative analysis of pixel domain processing and compressed domain processing.

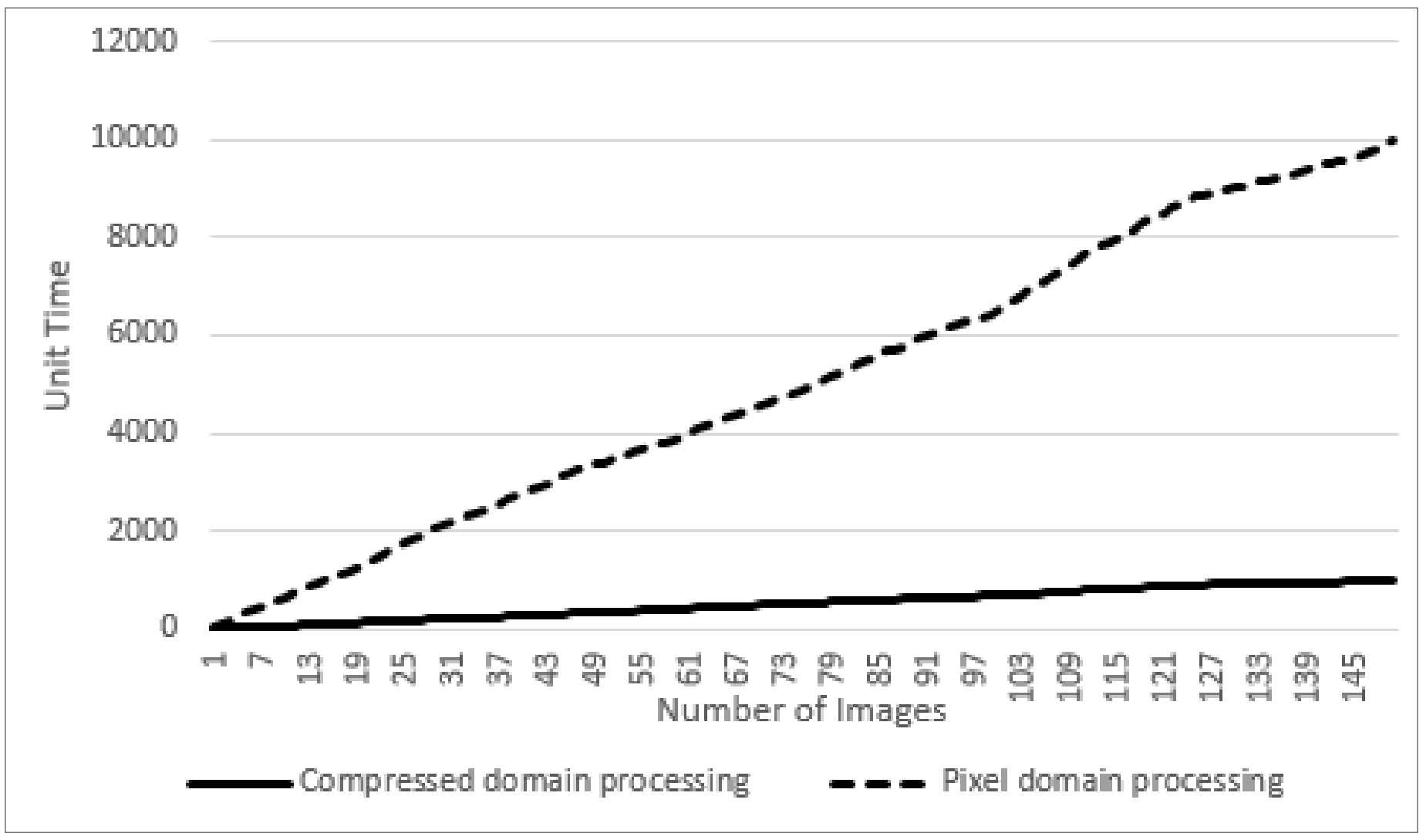

Figure 8: Comparative computational time analysis of CDP and PDP for ICDAR13 dataset

In the literature, authors of [Javed et al, 2016] have worked on compressed machine printed document images for word segmentation. Authors have shown promising results by varying font-styles and font-size in the documents. However, it is observed that the method cannot be directly applied to the compressed handwritten document images. Further, the method is specifically developed for the document images without skew, oscillation, etc.

## 2.2. Experimentation

Experiments are conducted on RLE compressed handwritten dataset namely ICDAR13, and PBOK (Section 1.6.1). The system is evaluated by counting the number of matches between the entities (words) detected by the algorithm and the entities in the ground truth [Nikolaos et al, 2013]. The Accuracy rate (AR) is defined as follows:

$$AR = \frac{Total\ number\ of\ one\ to\ one\ matching}{Total\ number\ of\ words} \times 100 \quad (7)$$

Table 2 shows the accuracy rate on evaluating the algorithms on the handwritten datasets. The word segmentation approach is mainly developed for compressed English handwritten document images. The experimentation is also conducted on the non-English handwritten document images to evaluate the performance of the model. It is observed that the efficacy is affected mainly because of uneven word spaces, and isolated components of the word.

Table 2: Accuracy rate when tested with various run-length compressed datasets

| Datasets (Handwritten) | Word segmentation accuracy rate (%) |
|---|---|
| ICDAR13 (English) | 90.29 |
| Kannada | 88.54 |
| Oriya | 83.86 |
| Bangla | 86.01 |
| Persia | 83.72 |

# 3. Character segmentation

In this section, we extend the previously discussed word segmentation technique to accomplish character segmentation specifically focusing on *English handwritten document images in the compressed format*. Here, the notion is to further analyse and refine every connected component in a text-line for locating the character spacing, which was established in the word segmentation approach. The primary step is to locate the Region of Interest (ROI) in a word by ignoring the ascenders and descenders of the characters, as they may overlap with the adjacent characters. Next step is to analyze the ROI for separating the characters. Here, ROI is horizontally divided into top, middle and bottom portions. Similar to the word segmentation approach, connected components are identified for every divided portion. Next, based on the length of the connected component, gaps between the adjacent characters are identified. This is merely based on the connectivity (or a number of transactions) between the adjacent characters. In other words, minimal edge(s) connecting the adjacent characters need to be identified and segmented. In English language, the connection between the adjacent characters mostly occurs in the mid-bottom regions. Therefore, with reference to the top portion, the characters are separated.

The procedure is illustrated with an example as shown in Figure 9, comprising a word of 8 characters. From the word segmentation approach, three connected components are obtained. Here, the character, ‘d’, does not overlap with its neighbouring characters as shown in Figure 10.

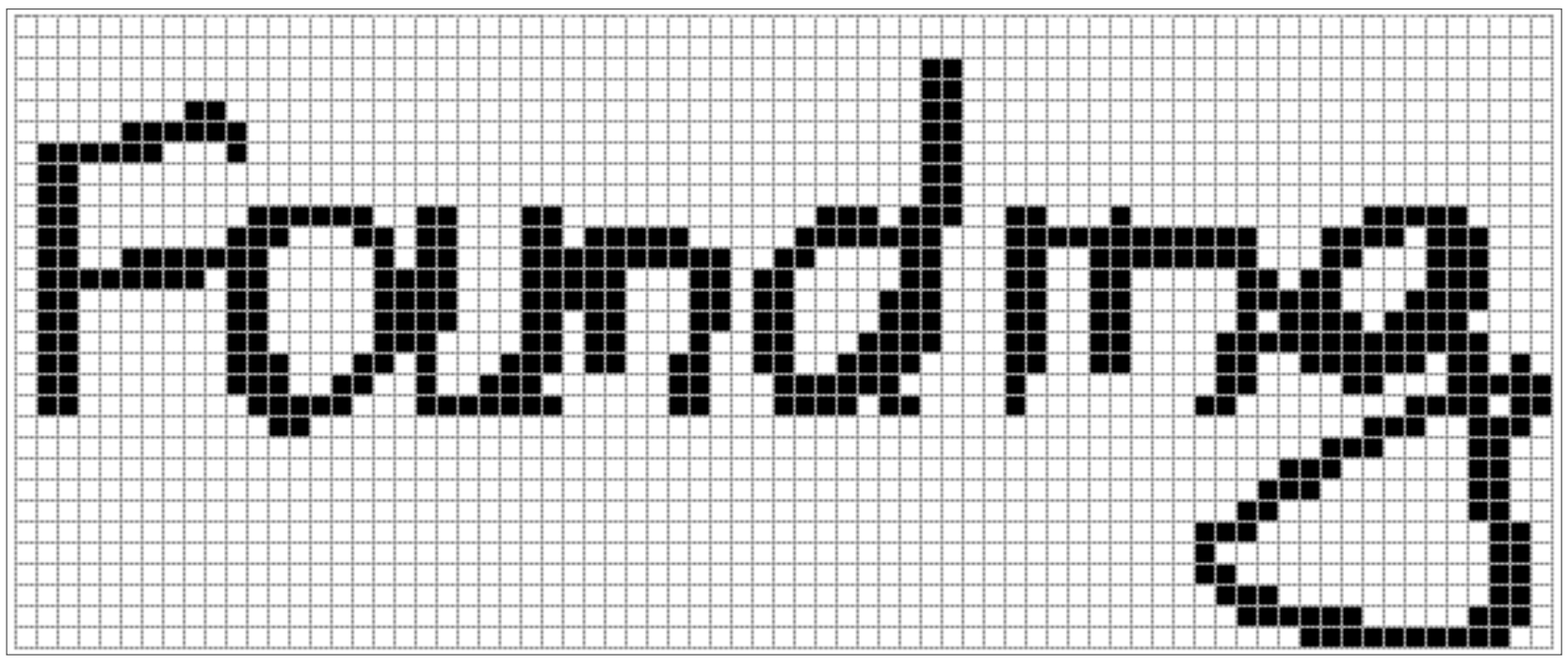

Figure 9: A sample word image

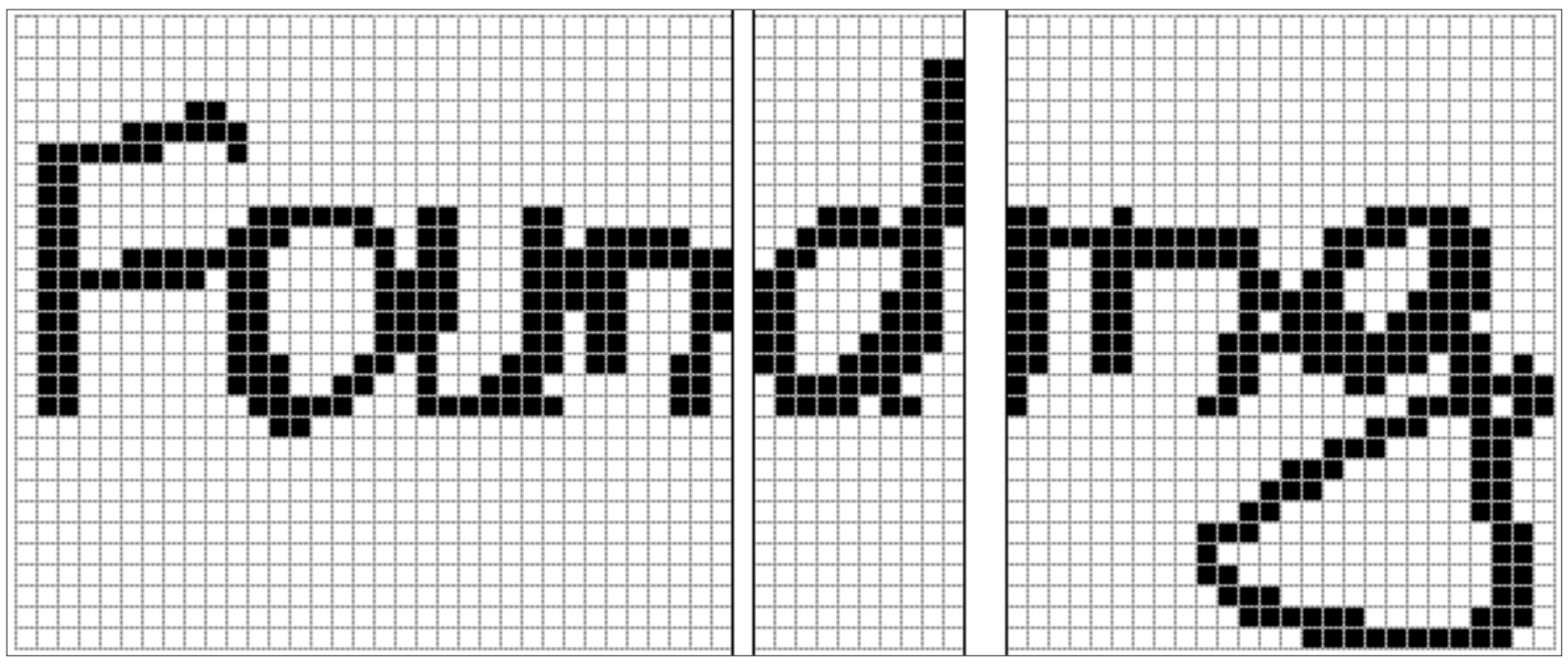

Figure 10: Segmented components

The next step is to find the region of interest (ROI) for analysis. Here, a threshold value is set empirically for the selection of ROI. Figure 11 depicts the ROI for the given example.

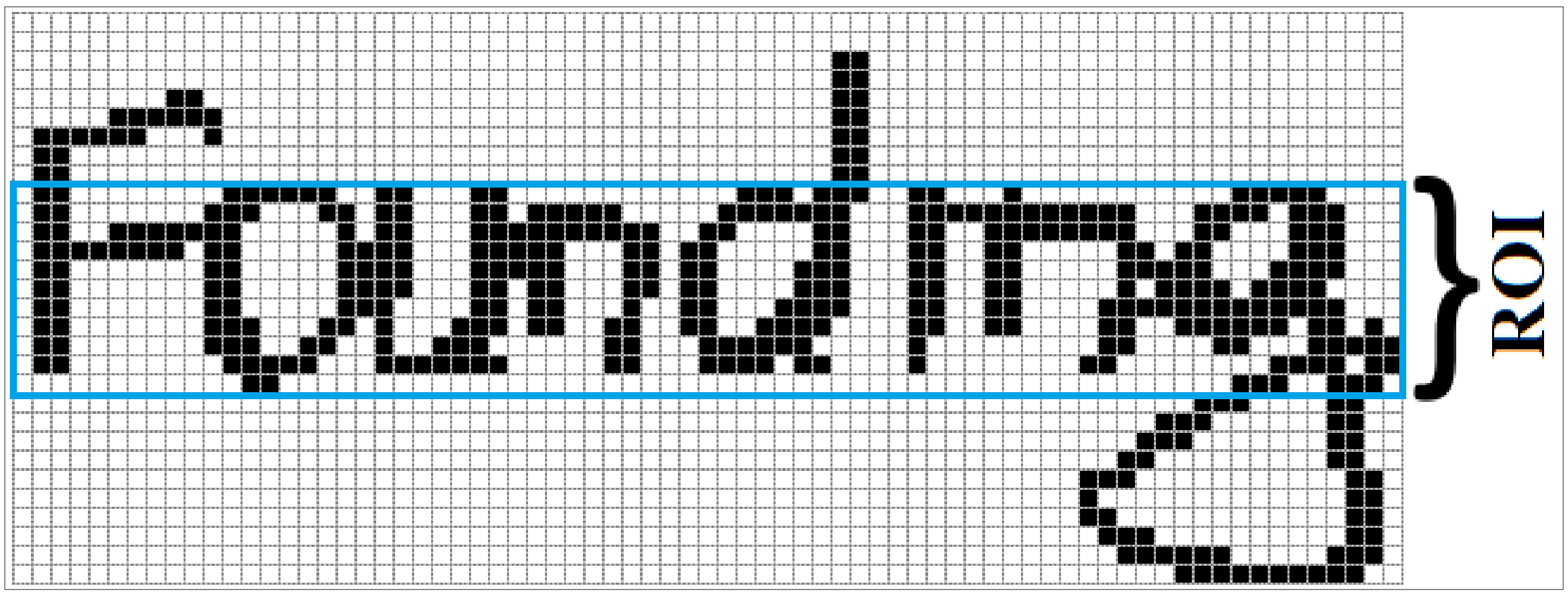

Figure 11: Region of interest for character separation

The next step is the division of ROI into three horizontal blocks. Then, the connected components are established for every portion, as we obtained in the word segmentation approach. It is observed that the overlapping of characters takes place in the middle portion. Therefore, a minimum cut-edge is identified with respective to the region for character segmentation. Figure 12 and 13 show the divided sections and the corresponding frequency of foreground runs.

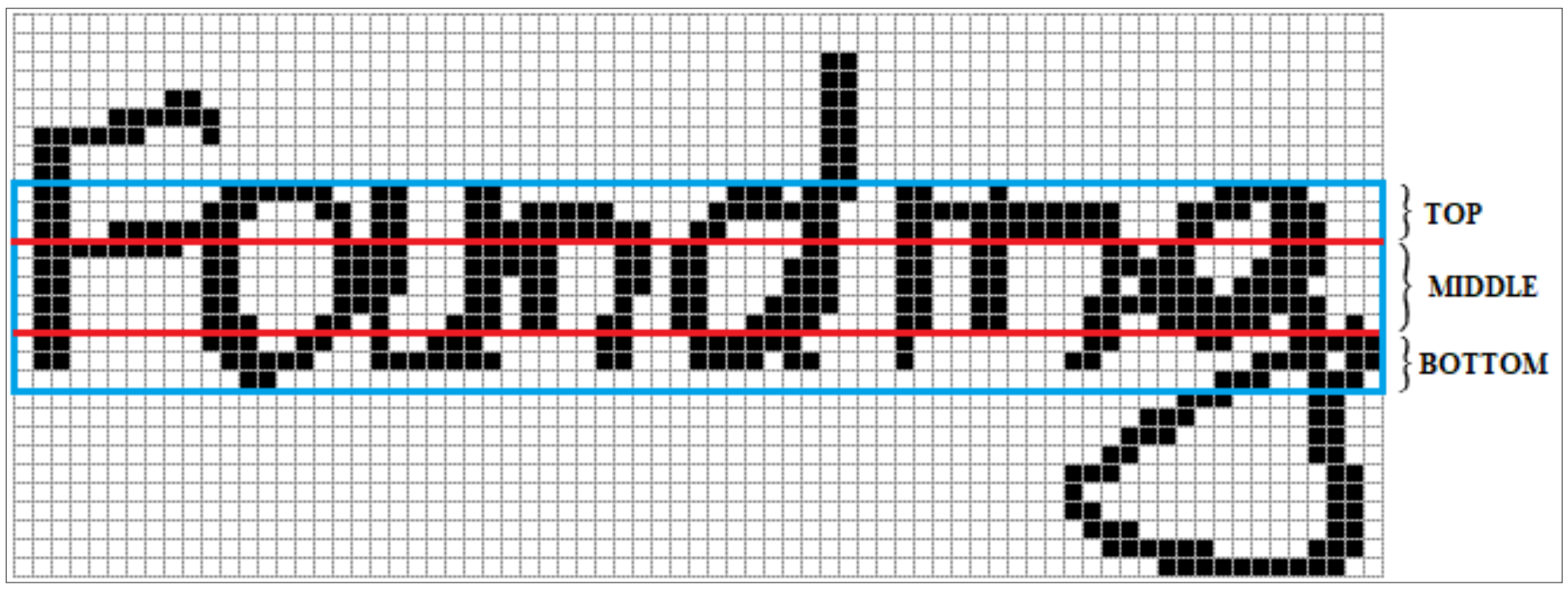

Figure 12: Divided portions for analysis

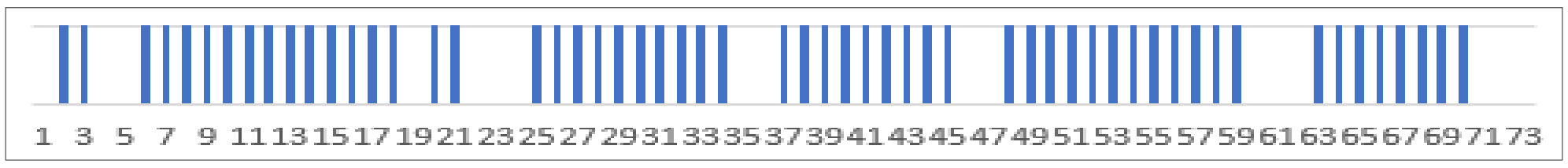

(a) Top portion

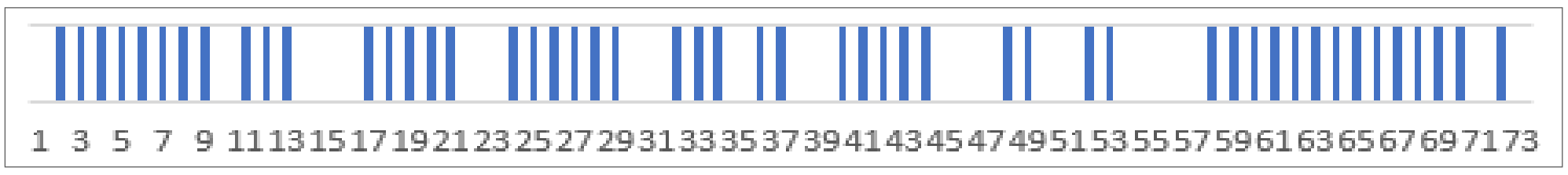

(b) Middle portion

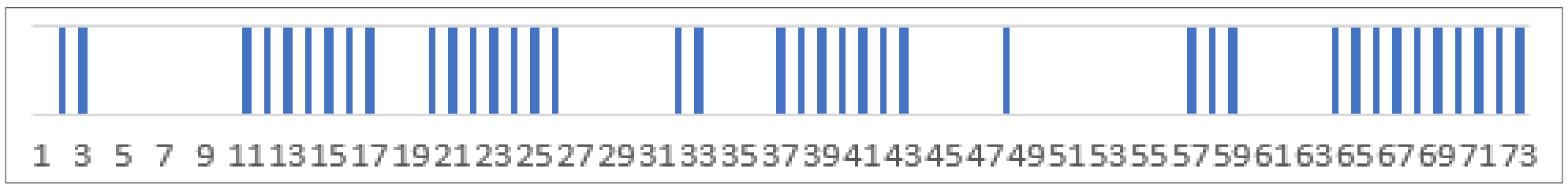

(c) Bottom portion

Figure 13: Frequency of bands in every divided portion

Next for finding a minimum cut edge, logical 'OR' operation is performed between the top and the bottom portions. Gaps between the frequency bands is considered as an ideal point (minimum cut edge) to separate the adjacent characters. Figure 14 (b) shows the separation of the characters.

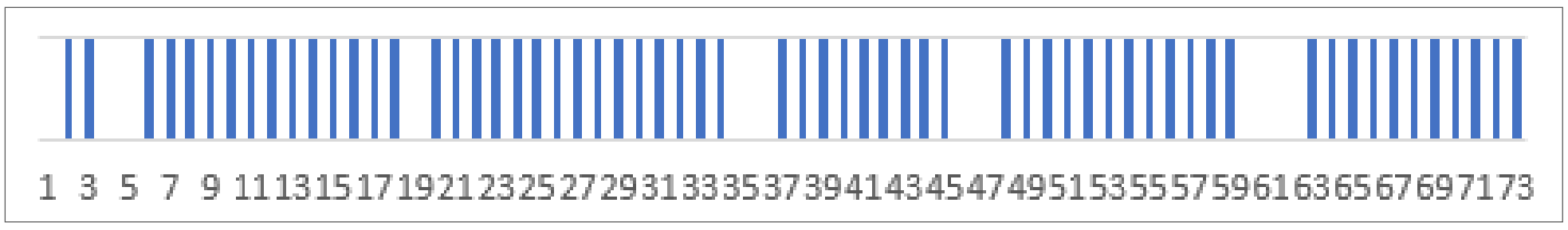

(a) OR operation of top and bottom portions

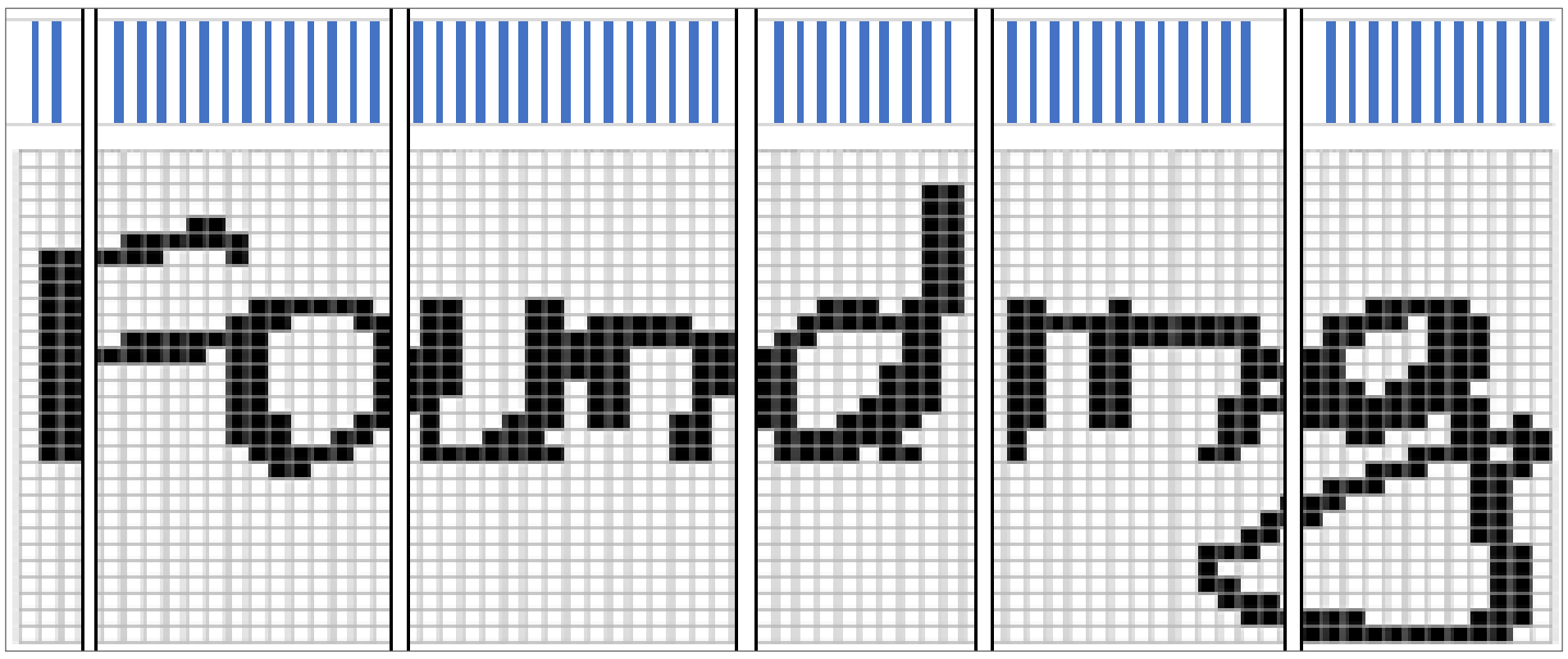

(b) Separated blocks

Figure 14: Character segmentation results.

The final step is to identify over-segmentation and under-segmentation regions. This is entirely based on the size of every connected component. Table 3 presents the statistics for the example. Here, the left and right margins are ignored as it attributes for word segmentation. It is also observed that there are 8 letters in the word. Length of the first component, c1, is negligible, therefore, the segmentation is assumed as an over-segmented region. If the length of a component is found to be larger than its average, then it is assumed as an under-segmentation region. Based on the frequency, the separator points are either added or removed. In this example, the separator points that divides component 1 and component 2 is removed. Figure 15 shows the final result after the insertion and deletion operations.

Table 3: Word segmented result for the example

| Components | Xmin | Xmax | Length |
|---|---|---|---|
| 1 | 2 | 3 | 2 |
| 2 | 6 | 18 | 13 |
| 3 | 20 | 34 | 15 |
| 4 | 37 | 45 | 9 |
| 5 | 48 | 59 | 12 |
| 6 | 63 | 73 | 11 |

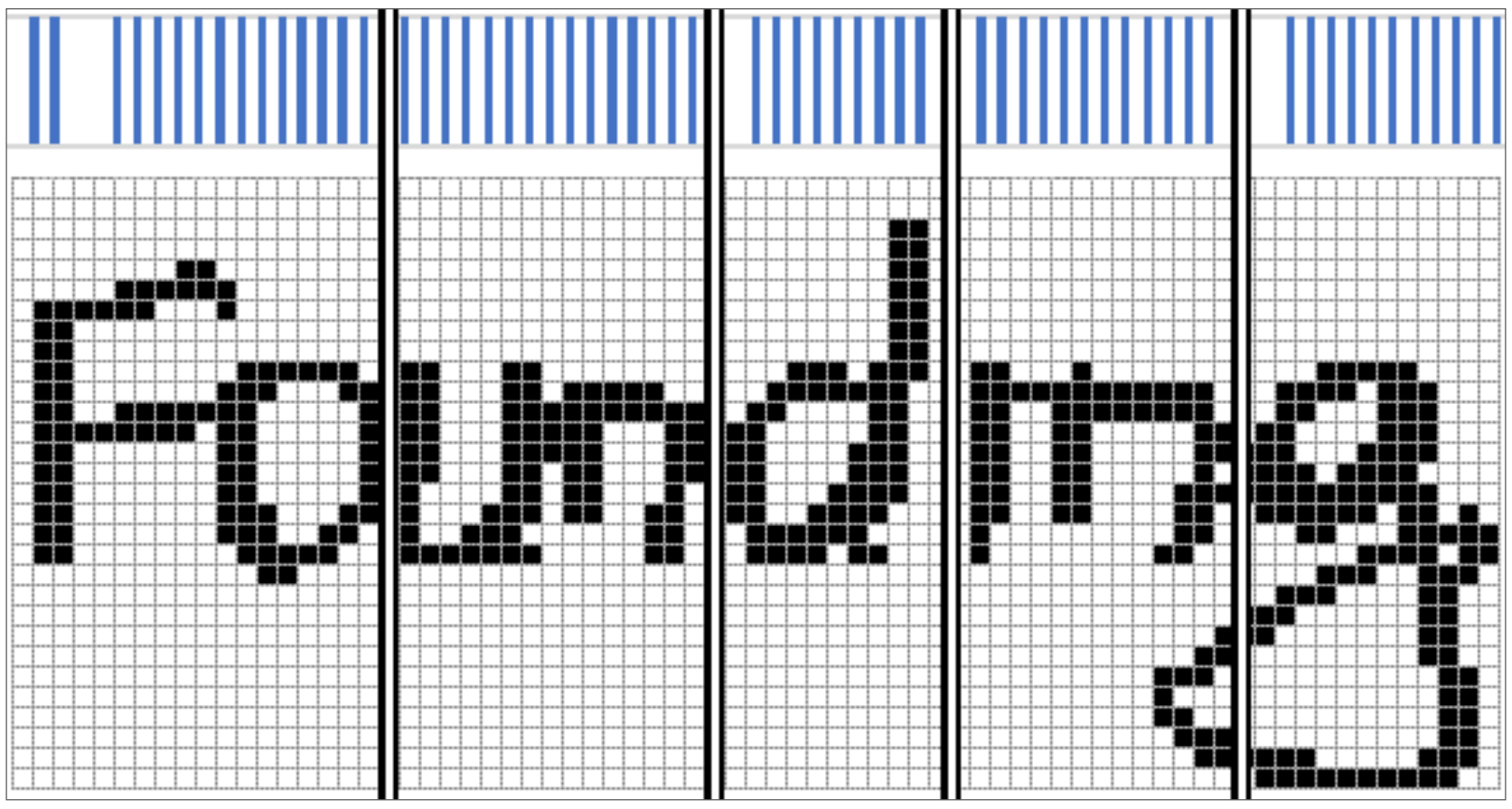

Figure 15: Result of character segmentation

Figure 16 shows the results of character segmentation in the uncompressed version.

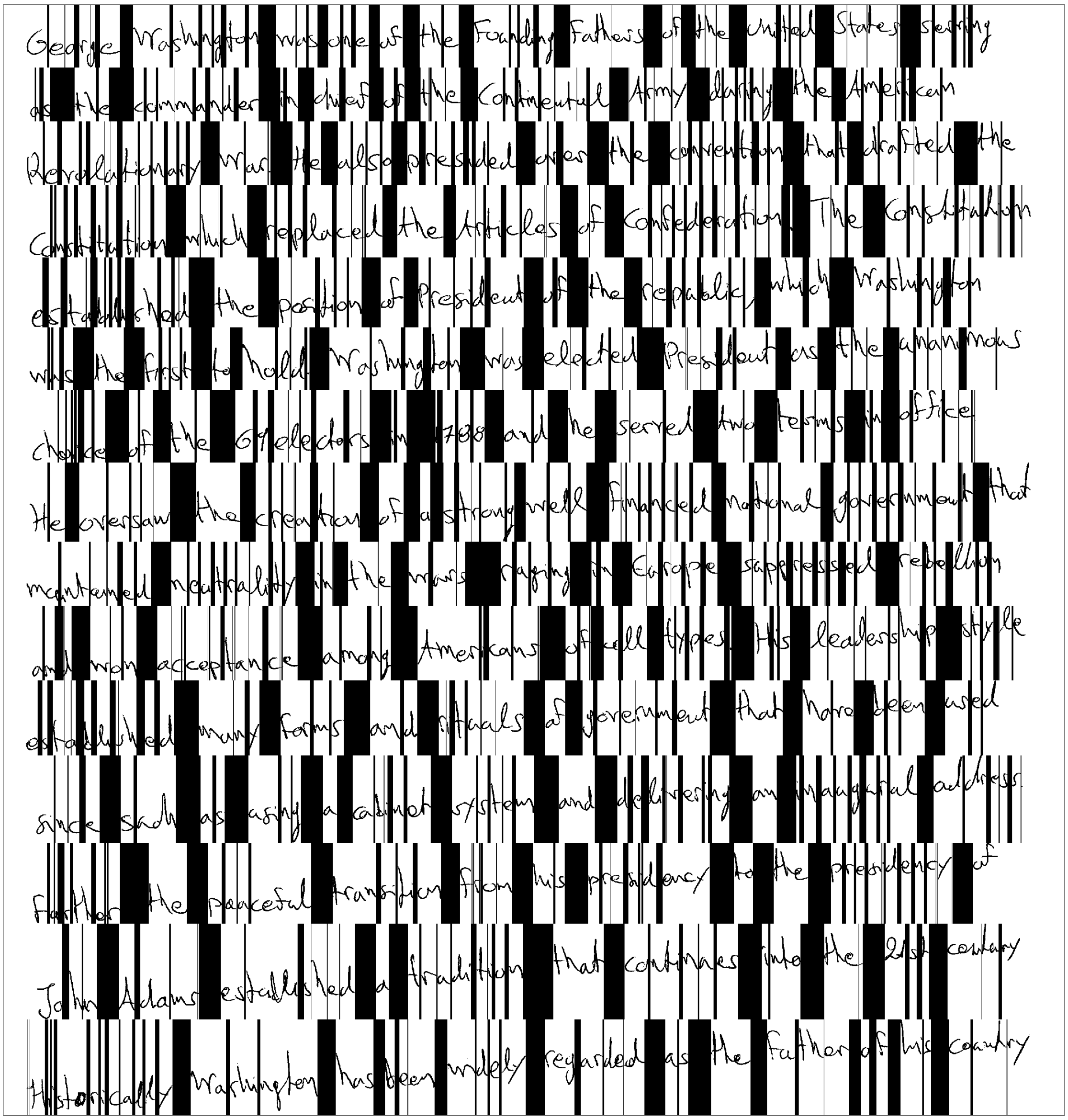

Figure 16: Results of character segmentation in the uncompressed version

## 3.1. Algorithm and Complexity Analysis

The inputs to the algorithm would be run-length (RLE) representation of the segmented word (Algorithm in Section 2.1), and a threshold value for ROI. The output would be the coordinate positions in RLE denoting character segmentation.

**Algorithm:** Character segmentation

***Input:*** *Word segments in RLE format, t (the threshold for ROI)*

***Output:*** *Coordinate Position (CP) in RLE*

| | |
|---|---|
| Step 1 | Find the ROI (middle region of the word) using t |
| Step 2 | Divide the ROI into 3 horizontal blocks |
| Step 3 | Find the connected components using equations (2, 3 and 4) for every block |
| Step 4 | Perform logical OR between Top and Bottom (Language dependent) |
| Step 5 | Identify the gaps in the pattern for character segmentation |
| Step 6 | Resolve over and under-segmentation problems based on lengths (of every component). |
| Step 7 | Identify the location in RLE using the spatial coordinates |
| Step 8 | Stop |

The height and width of the RLE are represented as m and n respectively. The time complexity of the character segmentation algorithm is $O(m \times n)$ in its worst-case scenario. If the same approach is employed in pixel domain, the best-case scenario would be $O(h \times w)$, where h and w represent the height and width of the uncompressed image. The comparative analysis is presented in Figure 17.

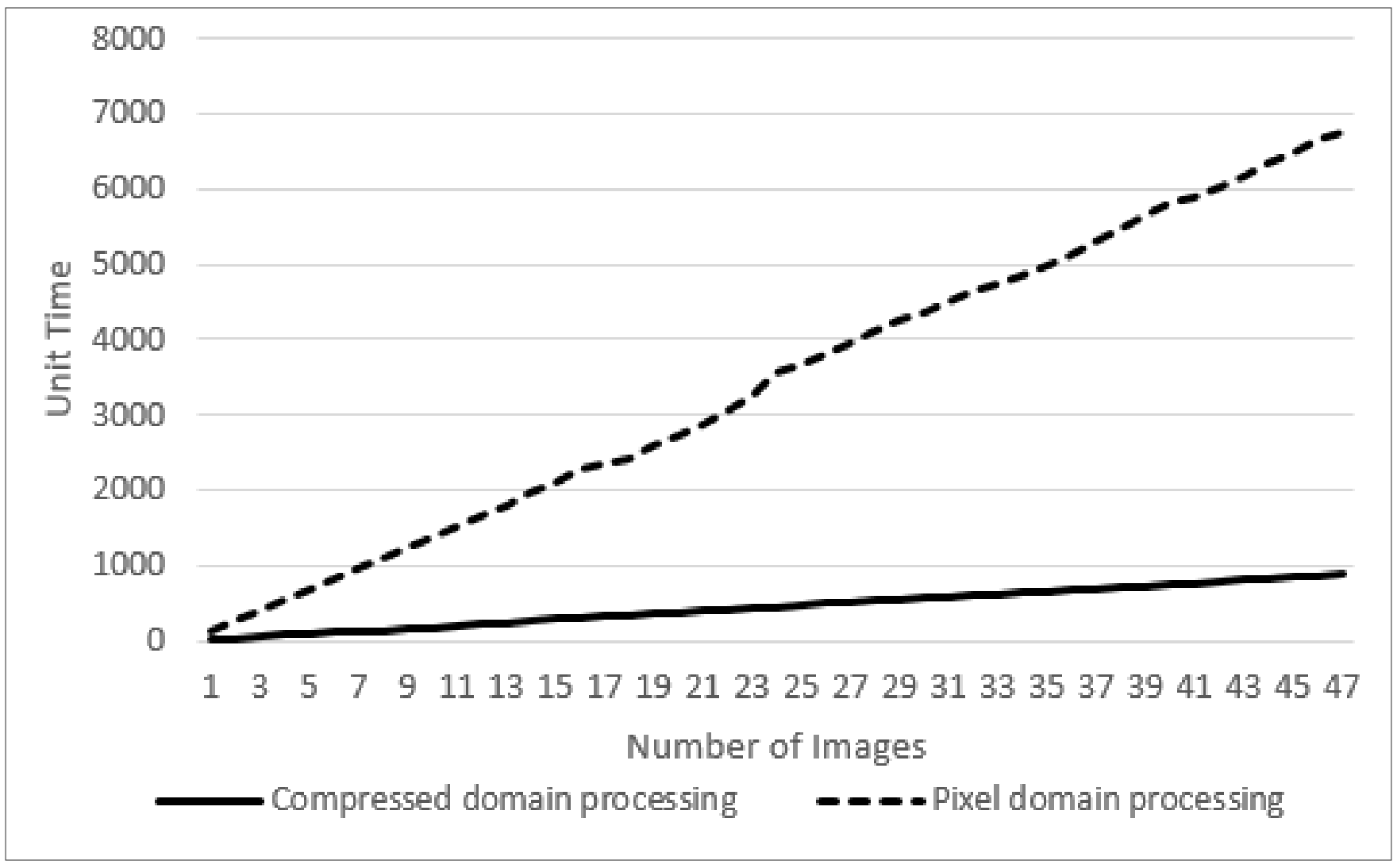

Figure 17: Comparative computational time analysis of CDP and PDP for ICDAR13 dataset

In the literature, authors of [Javed et al, 2016] have worked on compressed machine printed document images for character segmentation. Authors have shown promising results by varying font-styles and font-size in the documents. The work is based on the character spaces in the base-line. However, it is observed that the method cannot be directly applied directly to the compressed handwritten document images. Further, the method is specifically developed for the document images without skew, oscillation, etc.

## 3.2. Experimentation

Experiments are conducted on RLE compressed dataset namely ICDAR13, and PBOK (Section 1.6.1). The system is evaluated by counting the number of matches between the entities (text-lines) detected by the algorithm and the entities in the ground truth [Nikolaos et al, 2013]. The Accuracy rate (AR) is defined as follows:

$$AR = \frac{Total\ number\ of\ one\ to\ one\ matching}{Total\ number\ of\ characters} \times 100 \quad (8)$$

Table 4 shows the accuracy rate of character segmentation on testing it on the run-length compressed handwritten document images. The proposed model is also tested on other languages. The tabulated accuracy rate is the cascaded result of word and character segmentation approaches. The algorithm is specifically developed for English hand-scribed document images, therefore promising accuracy rate could be achieved. The same algorithm is also evaluated for the non-English scripts to check the performance of the model. As stated earlier (Section 2.2), accuracy rate is affected because of isolated words/characters, elongated ascenders and descenders of the characters and uneven word and character spaces in a text-line. These issues are more common with Persian and other Indic languages. Therefore, the accuracy rate for the non-English scripts are compromised based on the severity of the stated issues.

Table 4: Accuracy rate when tested with various run-length compressed dataset

| **Datasets (Handwritten)** | **Word Segmentation accuracy rate (%)** | **Character segmentation accuracy rate (%) (the cascaded result of Word segmentation)** |
|---|---|---|
| ICDAR13 (English) | 90.29 | 86.23 |
| Kannada | 88.54 | 83.98 |
| Oriya | 83.86 | 79.56 |
| Bangla | 86.01 | 79.31 |
| Persia | 83.72 | 75.32 |

# 4. Experimental results on Compressed Machine Printed Document Images

This section presents some of the experimental results of word segmentation and character segmentation that have been carried out on compressed machine printed document images. Two models are presented in Figure 18, and 19 respectively.

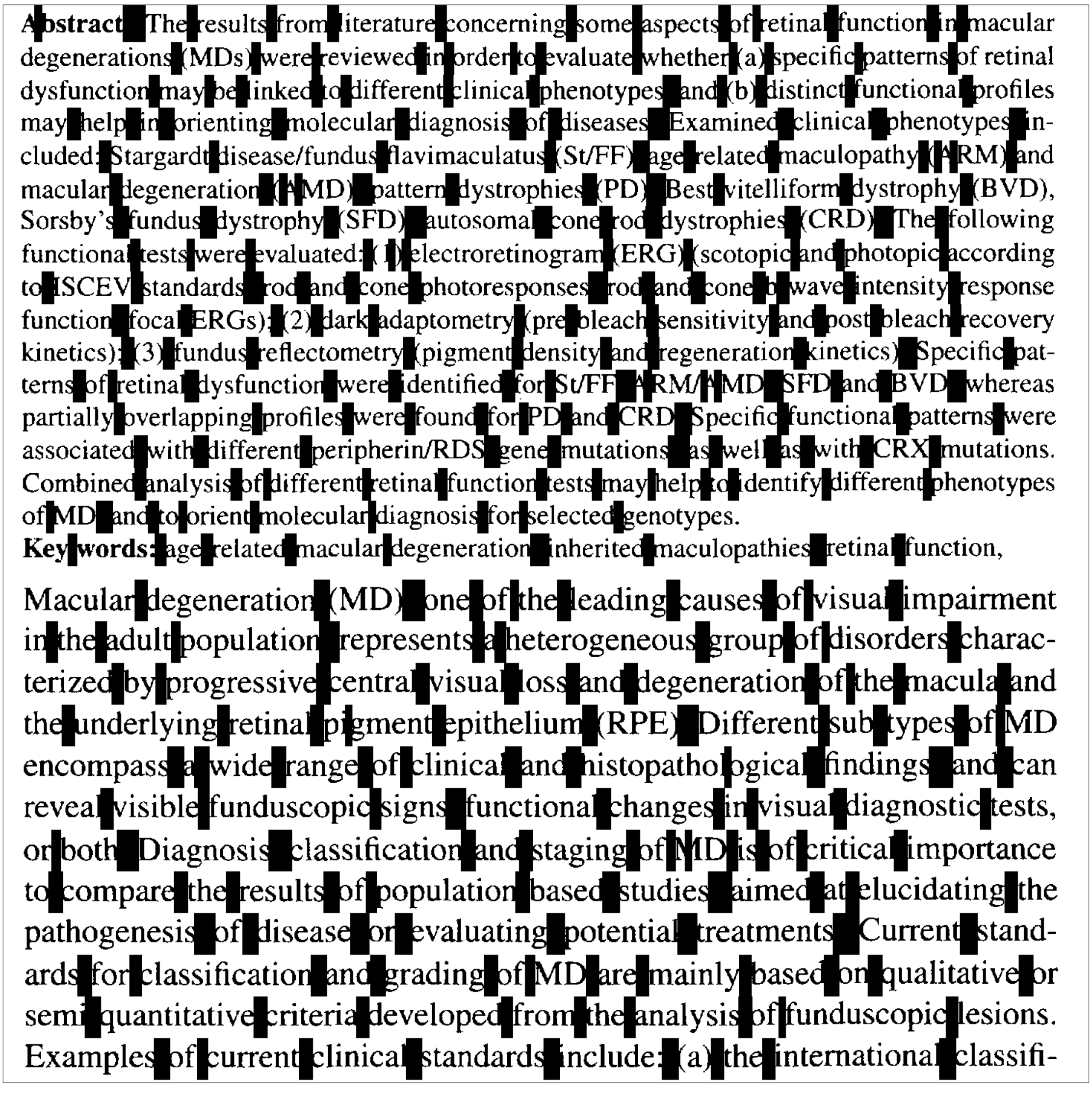

**Abstract** The results from literature concerning some aspects of retinal function in macular degenerations (MDs) were reviewed in order to evaluate whether (a) specific patterns of retinal dysfunction may be linked to different clinical phenotypes and (b) distinct functional profiles may help in orienting molecular diagnosis of diseases. Examined clinical phenotypes included: Stargardt disease/fundus flavimaculatus (St/FF), age related maculopathy (ARM) and macular degeneration (AMD), pattern dystrophies (PD), Best vitelliform dystrophy (BVD), Sorsby's fundus dystrophy (SFD), autosomal cone rod dystrophies (CRD). The following functional tests were evaluated: (1) electroretinogram (ERG) (scotopic and photopic according to ISCEV standards, rod and cone photoresponses, rod and cone b wave intensity response function, focal ERGs); (2) dark adaptometry (pre bleach sensitivity and post bleach recovery kinetics); (3) fundus reflectometry (pigment density and regeneration kinetics). Specific patterns of retinal dysfunction were identified for St/FF, ARM/AMD, SFD and BVD, whereas partially overlapping profiles were found for PD and CRD. Specific functional patterns were associated with different peripherin/RDS gene mutations as well as with CRX mutations. Combined analysis of different retinal function tests may help to identify different phenotypes of MD and to orient molecular diagnosis for selected genotypes.

**Key words:** age related macular degeneration, inherited maculopathies, retinal function,

Macular degeneration (MD), one of the leading causes of visual impairment in the adult population, represents a heterogeneous group of disorders characterized by progressive central visual loss and degeneration of the macula and the underlying retinal pigment epithelium (RPE). Different sub types of MD encompass a wide range of clinical and histopathological findings and can reveal visible funduscopic signs, functional changes in visual diagnostic tests, or both. Diagnosis, classification and staging of MD is of critical importance to compare the results of population based studies aimed at elucidating the pathogenesis of disease or evaluating potential treatments. Current standards for classification and grading of MD are mainly based on qualitative or semi quantitative criteria developed from the analysis of funduscopic lesions. Examples of current clinical standards include: (a) the international classifi-

Figure 18: Word Segmentation (Section 2): MARG dataset

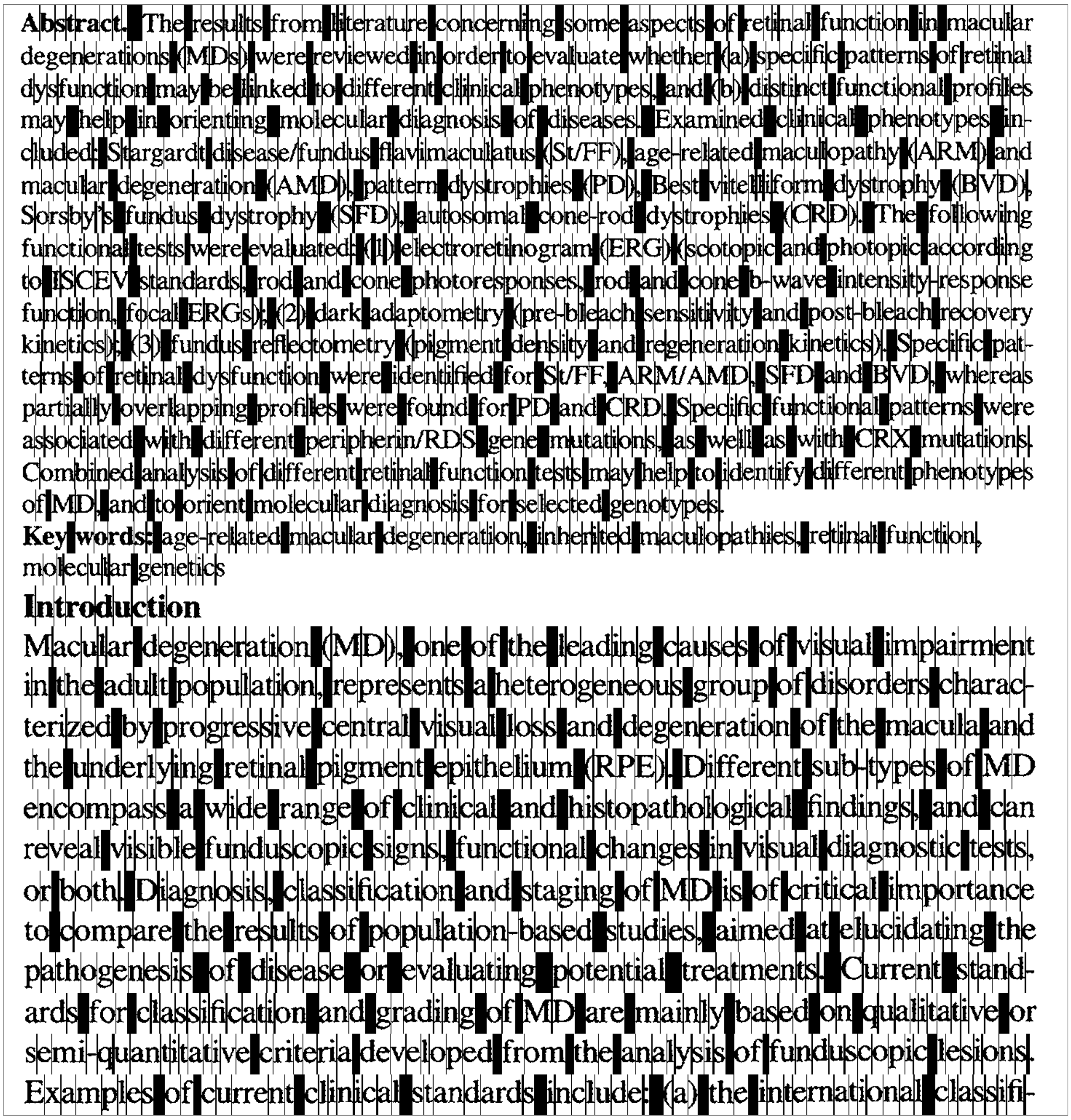
**Abstract.** The results from literature concerning some aspects of retinal function in macular degenerations (MDs) were reviewed in order to evaluate whether (a) specific patterns of retinal dysfunction may be linked to different clinical phenotypes, and (b) distinct functional profiles may help in orienting molecular diagnosis of diseases. Examined clinical phenotypes included: Stargardt disease/fundus flavimaculatus (St/FF), age-related maculopathy (ARM) and macular degeneration (AMD), pattern dystrophies (PD), Best vitelliform dystrophy (BVD), Sorsby's fundus dystrophy (SFD), autosomal cone-rod dystrophies (CRD). The following functional tests were evaluated: (1) electroretinogram (ERG) (scotopic and photopic according to ISCEV standards, rod and cone photoresponses, rod and cone b-wave intensity-response function, focal ERGs); (2) dark adaptometry (pre-bleach sensitivity and post-bleach recovery kinetics); (3) fundus reflectometry (pigment density and regeneration kinetics). Specific patterns of retinal dysfunction were identified for St/FF, ARM/AMD, SFD and BVD, whereas partially overlapping profiles were found for PD and CRD. Specific functional patterns were associated with different peripherin/RDS gene mutations, as well as with CRX mutations. Combined analysis of different retinal function tests may help to identify different phenotypes of MD, and to orient molecular diagnosis for selected genotypes.
**Key words:** age-related macular degeneration, inherited maculopathies, retinal function, molecular genetics
**Introduction**
Macular degeneration (MD), one of the leading causes of visual impairment in the adult population, represents a heterogeneous group of disorders characterized by progressive central visual loss and degeneration of the macula and the underlying retinal pigment epithelium (RPE). Different sub-types of MD encompass a wide range of clinical and histopathological findings, and can reveal visible funduscopic signs, functional changes in visual diagnostic tests, or both. Diagnosis, classification and staging of MD is of critical importance to compare the results of population-based studies, aimed at elucidating the pathogenesis of disease or evaluating potential treatments. Current standards for classification and grading of MD are mainly based on qualitative or semi-quantitative criteria developed from the analysis of funduscopic lesions. Examples of current clinical standards include: (a) the international classifi-

Figure 19: Character Segmentation (Section 3): MARG dataset

The results show that the efficacy of the proposed models on machine printed document images is found to be very high when compared to the handwritten document images. In the literature [Javed et al (Ph.D. Thesis), 2016], authors have worked on the machine printed document images for word and character segmentation. However, the models have been demonstrated on the documents which are completely free from skew.

## 5. Conclusion

In this article, we have proposed techniques for segmenting the compressed handwritten document images into words and characters. The word segmentation approach is based on the inter-word gaps whereas the character segmentation is based on the intra-word spacing. We have employed graph-based min-cut concept for segmenting the characters. Experimentations have been carried out on the run-length compressed handwritten document images. In this article, methods have been developed specifically focusing on *Compressed English handwritten document images*. However, the algorithms are evaluated on *Compressed Non-English documents* to check the performance of the proposed model. Further, to demonstrate the significance of the proposal models, the algorithms have been experimented on compressed machine printed document images and the results are shown in Section 4. Comparative analysis of pixel domain and compressed domain is presented to demonstrate the efficacy of the system. The developed models can withstand skew up-to certain level, however, this issue is being kept beyond the scope of the present work.